\documentclass[10pt,twocolumn,letterpaper]{article}

\usepackage{cvpr}              % To produce the CAMERA-READY version
\usepackage{graphicx}
\usepackage{amsmath}
\usepackage{amssymb}
\usepackage{booktabs}

\usepackage{bbm}
\usepackage{multirow}
\usepackage{enumitem}
\usepackage[ruled]{algorithm2e}
\usepackage[dvipsnames]{xcolor}
\usepackage{array}
\usepackage[accsupp]{axessibility}

\usepackage[pagebackref,breaklinks,colorlinks]{hyperref}

\DeclareMathOperator*{\argmax}{arg\,max}

\usepackage{color}

\def\Approach{iFS-RCNN}

\def\block1{{$\sf block1$}}
\def\block2{{$\sf block2$}}
\def\block3{{$\sf block3$}}
\def\block4{{$\sf block4$}}
  
\def\Real{{\mathbb R}}

\newcolumntype{P}[1]{>{\centering\arraybackslash}p{#1}}

\usepackage[capitalize]{cleveref}
\crefname{section}{Sec.}{Secs.}
\Crefname{section}{Section}{Sections}
\Crefname{table}{Table}{Tables}
\crefname{table}{Tab.}{Tabs.}

\def\cvprPaperID{5432} % *** Enter the CVPR Paper ID here
\def\confName{CVPR}
\def\confYear{2022}

\begin{document}

%%%%%%%%% TITLE - PLEASE UPDATE
\title{iFS-RCNN: An Incremental Few-shot Instance Segmenter}

\author{Khoi Nguyen \\
VinAI Research,
Ha Noi, Vietnam \\
{\tt\small ducminhkhoi@gmail.com}
% For a paper whose authors are all at the same institution,
% omit the following lines up until the closing ``}''.
% Additional authors and addresses can be added with ``\and'',
% just like the second author.
% To save space, use either the email address or home page, not both
\and
Sinisa Todorovic \\
Oregon State University,
Oregon, USA\\
{\tt\small sinisa@oregonstate.edu}
}
\maketitle

%%%%%%%%% ABSTRACT
\begin{abstract}
   This paper addresses incremental few-shot instance segmentation, where a few examples of new object classes arrive when access to training examples of old classes is not available anymore, and the goal is to perform well on both old and new classes. We make two contributions by extending the common Mask-RCNN framework in its second stage -- namely, we specify a new object class classifier based on the probit function and a new uncertainty-guided bounding-box predictor. The former leverages Bayesian learning to address a paucity of training examples of new classes. The latter learns not only to predict object bounding boxes but also to estimate the uncertainty of the prediction as a guidance for bounding box refinement. We also specify two new loss functions in terms of the estimated object-class distribution and bounding-box uncertainty. Our contributions produce significant performance gains on the COCO dataset over the state of the art -- specifically, the gain of +6 on the new classes and +16 on the old classes in the AP instance segmentation metric. Furthermore, we are the first to evaluate the incremental few-shot setting on the more challenging LVIS dataset.
\end{abstract}

%%%%%%%%% BODY TEXT
\section{Introduction}
\label{sec:introduction}

\begin{figure}[h!]
    \centering
    \includegraphics[scale=0.35]{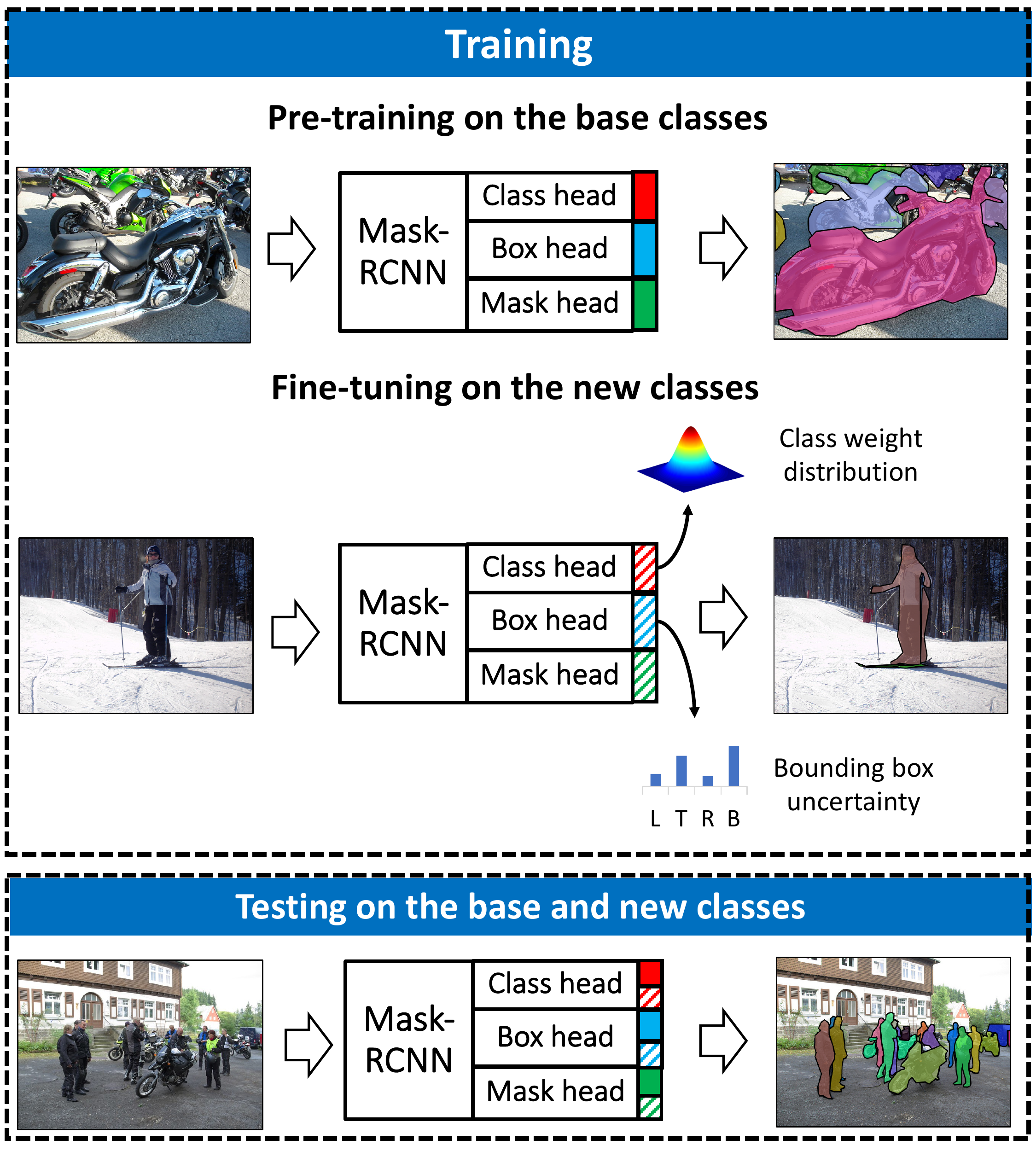}
    % \vspace{-10pt}
    \caption{Our \Approach~is first pre-trained on abundant examples of base classes, and then fine-tuned on a few examples of new classes.  \Approach~modifies the  classification head of Mask-RCNN by estimating the class-weight distribution via Bayesian learning. \Approach~modifies the  bounding-box head of Mask-RCNN
    by computing uncertainty of  predicting the left (L), right (R), top (T), and bottom (B) sides of the bounding boxes. For testing, the last layers learned on the new classes (diagonal stripes) are concatenated with the corresponding ones learned on the base classes  (solid color).}
    \label{fig:problem_statement}
    \vspace{-10pt}
\end{figure}

This paper addresses the two related problems of incremental few-shot object detection (iFSOD) and instance segmentation (iFSIS). Initially, we are given a large training set of base object classes, which can be used for pre-training an instance segmenter. After this pre-training, access to training examples of the base classes becomes unavailable. With an arrival of a few training examples of new classes, the goal is to achieve successful object detection and instance segmentation on both new and base classes. Our key challenges include: how to address a paucity of data for new classes, and how to train on the new classes such that the base classes are not ``forgotten''. 

Tab.~\ref{tab:settings} compares iFSOD and iFSIS with other related problems. iFSOD and iFSIS are important problems arising in many applications, where access to old training data becomes unavailable, due to, e.g., privacy and security issues or new legal regulations of data access. Also, they are critical in applications where limited time budgets prohibit retraining on both base and new classes.

\begin{table}
\small
\begin{center}
\setlength{\tabcolsep}{2pt}
\begin{tabular}{|c|c|c|c|}
\hline
Settings & Pretrained on & Fine-tuned on & Tested on \\
\hline\hline
FSOD - FSIS & \textcolor{blue}{base} & \textcolor{red}{new} & new \\
gFSOD - gFSIS & \textcolor{blue}{base} & \textcolor{red}{base + new} & base + new \\
CL    & \textcolor{blue}{base} & \textcolor{blue}{new} & base + new \\
iFSOD - iFSIS & \textcolor{blue}{base} & \textcolor{red}{new} & base + new \\
\hline
\end{tabular}
\end{center}

\vspace{-10pt}
\caption{A comparison of related problems. The \textcolor{blue}{blue} and \textcolor{red}{red} indicate abundant and a few examples, respectively, of base and new classes. FSOD (FSIS): few-shot object detection (instance segmentation), gFSOD (gFSIS): generalized FSOD (FSIS), CL: continual learning, iFSOD (iFSIS): incremental FSOD (FSIS). iFSOD (iFSIS) are more challenging than: FSOD (FSIS), since we test on both classes; gFSOD (gFSIS), since our training cannot access the base classes; CL, since they work with more examples.}
\label{tab:settings}

\vspace{-10pt}
\end{table}

There is scant work on iFSOD and iFSIS. Following recent FSIS approaches, we use Mask-RCNN \cite{he2017mask}, and modify its prediction heads, as shown in Fig.~\ref{fig:problem_statement}. Mask-RCNN is first pre-trained with abundant examples of base classes, and then fine-tuned on new classes by ``freezing'' all modules except for the  classification head, bounding-box  head, and segmentation-mask head. Finally, for testing on both base and new classes,  weights learned on the new classes are concatenated with weights learned on the base classes to make the corresponding last layers in the classification, bounding-box, and segmentation-mask heads.

\begin{figure*}[h!]
    \centering
    \includegraphics[scale=0.5]{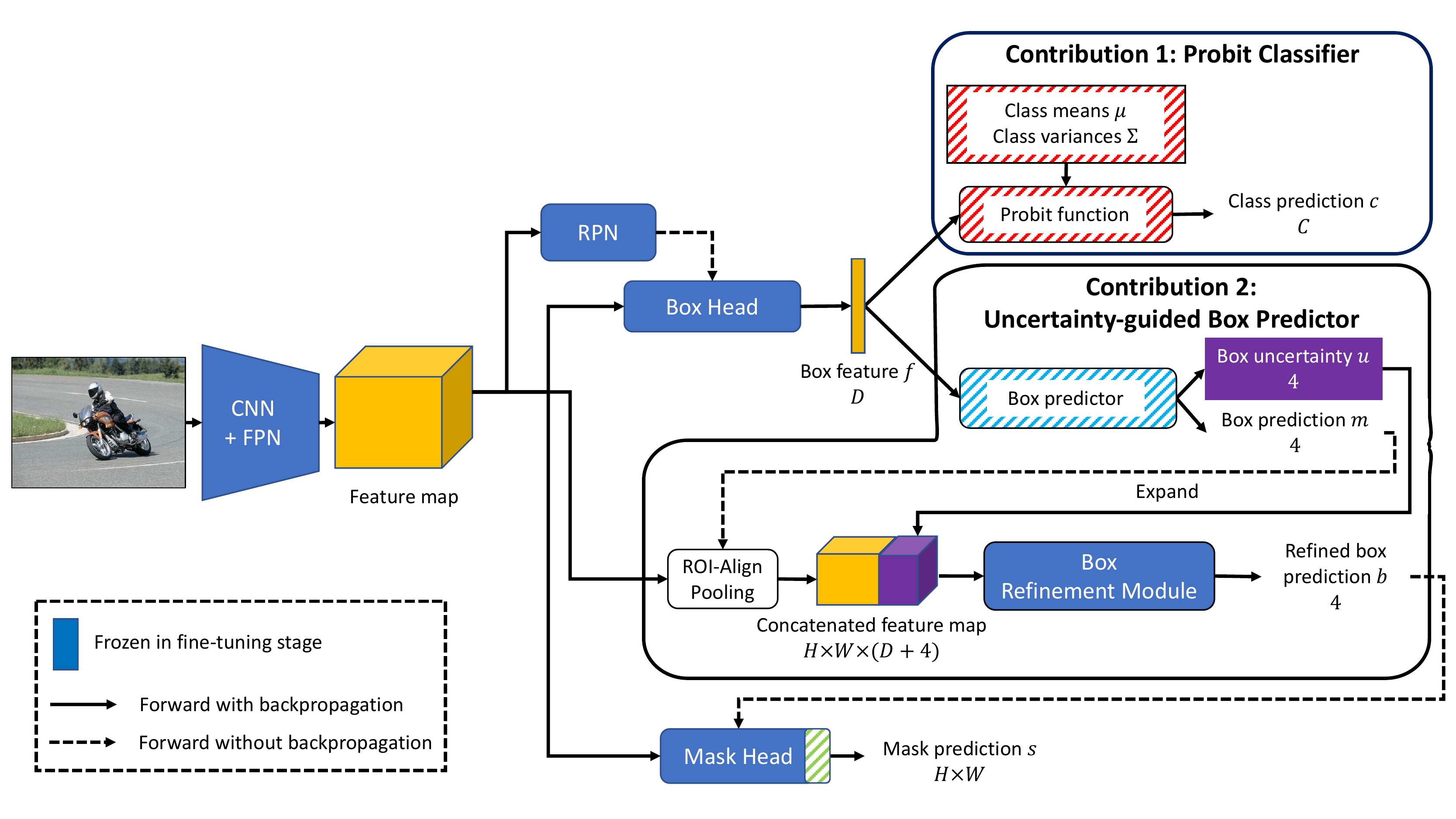}
    % \vspace{-10pt}
    \caption{\Approach~extends Mask-RCNN with two contributions: probit classifier and uncertainty-guided box predictor. The former uses Bayesian learning to estimate a probability distribution of the classifier head's weights  (red diagonal stripes). Our contribution 1 is the efficient, analytical formulation of this Bayesian learning using the probit function. The latter explicitly estimates uncertainty of predicting bounding boxes (blue diagonal stripes), and uses the estimated uncertainty (violet) as input features along with the ROI-align-pooled features (dark yellow) for refining the boxes. The colored stripes depict the last layers of the classification, box, and segmentation heads learned in the few-shot setting during fine-tuning of the new classes. }
    \label{fig:main_diagram}
    \vspace{-5pt}
\end{figure*}

As shown in Fig.~\ref{fig:problem_statement} and depicted in more detail in Fig.~\ref{fig:main_diagram}, we make two contributions aimed at addressing overfitting of Mask-RCNN in few-shot fine-tuning and improving its generalization to query images with large appearance-shape-scale variations of both base and new classes.  

Inspired by deep Bayesian learning \cite{blundell2015weight}, {\em our first contribution} is about learning a distribution of the classification head's weights on the new classes, and using the estimated distribution for regularization of the fine-tuning. Instead of using the standard Monte Carlo sampling of weights for this Bayesian learning, our key technical novelty is in casting the weight-distribution learning as a Bayesian logistic regression problem, and specifying an efficient approximation to this intractable problem using the probit function. From our ablation studies, our probit-based approximation gives a significantly better performance than the Monte Carlo sampling, even under a reasonable training-time budget.  

{\em Our second contribution} is about estimating the uncertainty of bounding-box localization on the new classes, and using the estimated uncertainty for two purposes -- to refine the bounding-box prediction and to appropriately weight the loss of bounding-box prediction. As shown in Fig.~\ref{fig:main_diagram}, we use the estimated uncertainty along with the ROI-aligned-pooled feature map as input to a new bounding-box refinement module. The refined bounding box is subsequently input to the segmentation head. Also, we define a new loss between the ground truth and predicted bounding box, such that the {\em loss becomes smaller for highly uncertain predictions}, i.e., our fine-tuning stronger penalizes errors on training examples with highly certain bounding-box predictions.

It is worth noting that we neither use Bayesian learning nor explicitly estimate uncertainty for fine-tuning of the segmentation head of Mask-RCNN. This is because fine-tuning of the segmentation head on a few examples of the new classes does not face the common challenges of few-shot learning. Recall that the segmentation head predicts pixel labels independently with a $1\times 1$ convolution. Consequently, every (pixel, label) pair is an independent training example, giving rise to a sufficiently large training set for fine-tuning of the segmentation mask.

% Our extension of Mask-RCNN with the probit-based Bayesian learning and estimation of bounding-box uncertainty gives the name to our approach -- \Approach. 
Our extension of Mask-RCNN for incremental few-shot setting gives the name to our approach -- \Approach. 
\Approach~is evaluated on the COCO dataset \cite{lin2014microsoft} for few-shot object detection and instance segmentation with the iFSOD, iFSIS, FSOD, and FSIS tasks. \Approach~significantly outperforms a recent approach \cite{Ganea_2021_CVPR}. In comparison with the standard Mask-RCNN trained in the gFSOD and gFSIS settings, \Approach~shows a considerable improvement on the new classes while retaining the same performance on the base classes. On iFSOD and iFSIS, we also achieve the higher COCO AP rates by +6 and +16 for the new and base classes, respectively, relative to the state of the art.

In addition, we are the first to report the results of iFSOD, FSIS, and iFSIS on the more challenging LVIS dataset \cite{gupta2019lvis} having significantly more classes and long-tailed class distributions. 

%In sum, we make the following two contributions:
%\begin{itemize}
%    \vspace{-2mm}
%    \item We introduce a new probit classifier that analytically learns a weight distribution of classification head addressing high variation in object appearance and scales. Also, the probit function uses the sigmoid function for approximation, it results in independent score predictions for each class directly addressing the iFSIS problem. This contribution is illustrated in the yellow box of Fig.~\ref{fig:main_diagram}.
%    \vspace{-2mm}
%    \item We propose a new uncertainty-guided box predictor which takes into account the uncertainty due to occlusions or similar appearance between objects when predicting bounding boxes under a few-shot setting. This contribution is illustrated in the red box of Fig.~\ref{fig:main_diagram}.
%\end{itemize}
%\vspace{-2mm}
In the following, Sec.~\ref{sec:related_work} reviews prior work; Sec.~\ref{sec:approach} specifies \Approach; and Sec.~\ref{sec:experiments} presents our implementation details and experimental results.

%-------------------------------------------------------------------------
\section{Related Work}
\label{sec:related_work}

This section reviews closely related work. 

% \textbf{Instance segmentation} approaches can be broadly divided into bounding-box-based and bounding-box-free. The former group \cite{he2017mask,li2017fully,chen2018masklab,liu2018path,bolya2019yolact} first detects bounding boxes, and then segments foreground pixels inside the boxes. The latter group \cite{kirillov2017instancecut,kong2018recurrent,chandra2017dense,fathi2017semantic,NEURIPS2020_3341f6f0} first performs semantic segmentation, and then groups pixels into instances. These approaches, however, are not suitable for iFSIS as they require a lot of training annotations and complete retraining on new classes.

\textbf{FSOD} approaches \cite{chen2018lstd,karlinsky2019repmet,kang2019few,yanICCV19metarcnn,wang2020frustratingly,fan2021generalized,wu2020multi,fan2020few,xiao2020few,li2021transformation,zhang2021meta} typically adapt Faster-RCNN \cite{ren2015faster}, YOLO \cite{redmon2018yolov3}, or DETR \cite{carion2020end} for standard object detection to few-shot setting. These approaches can be divided into two subgroups based on: episodic training \cite{kang2019few,yanICCV19metarcnn,zhang2021meta,xiao2020few,fan2020few} and fine-tuning \cite{chen2018lstd,wang2020frustratingly,wu2020multi,fan2021generalized}. The former uses episodic training to mimic the setting of few-shot learning, by limiting access to a few annotated support images for each base class. The latter fine-tunes the weights of some layers while freezing the rest to preserve knowledge learned on the base classes.

\textbf{FSIS} approaches typically use Mask-RCNN \cite{he2017mask} as their backbone network. As in FSOD, these methods either train episodically \cite{yan2019meta,michaelis2018oneshot,fan2020fgn,nguyen2021fapis} or pretrain on the base classes then fine-tune the last layers of each head \cite{wang2020frustratingly} on the new classes. 
% On the other hand, a recent fine-tuning approach \cite{wang2020frustratingly} demonstrated a better performance on few-shot object detection.
% The latter first pre-train Faster-RCNN with abundant examples of the base classes, and then fine-tune only the last layer of the classification and box heads on the new classes while keeping the other layers of Faster-RCNN unchanged. 
Our \Approach~follows the second training strategy and applies it to all three heads -- the classification, bounding-box, and segmentation heads --  of Mask-RCNN.  

\textbf{iFSOD} approaches \cite{perez2020incremental,li2021towards,li2021class} typically use Faster-RCNN as their backbone network. In \cite{li2021class}, knowledge distillation ensures the prediction of the base classes, after fine-tuning on the new classes, to match its pretrained prediction. In \cite{perez2020incremental}, weights of the box head are generated on the fly, based on a class-specific code extracted from examples of the target classes. In this way, each class has a distinct box head for object detection. Our \Approach~also uses for each class a distinct set of the classification, bounding-box, and segmentation heads of Mask-RCNN, where the difference between the respective heads is in the last layer fine-tuned for each new class separately.

%\textbf{Incremental few-shot semantic segmentation (iFSS)} is similar to iFSIS except that the goal is rather semantic than instance segmentation. A recent approach \cite{cermelli2020few} pre-trains their DeepLabv3 on base classes, and then fine-tunes the model on new classes with a cosine-similarity classifier. Also, a Batch Renormalization layer and knowledge distillation are introduced to alleviate ``forgetting''. 

\textbf{iFSIS}: A recent approach to iFSIS \cite{Ganea_2021_CVPR} replaces the standard fully-connected classifier in Mask-RCNN with the cosine-similarity classifier. Unlike our \Approach, they do not use Bayesian learning and do not estimate bounding-box uncertainty. They convert activations of the classification head to a softmax distribution, and train the head with cross-entropy loss. In this way, the activation scores of all classes compete with each other to determine the class of the bounding box. As training examples of the new classes are scarce, their activation scores are likely to be smaller than those of the base classes in the softmax function. Consequently, their classifier is likely to be biased toward favoring the base classes. They address this bias by using the cosine-similarity classifier, where both the box feature and class weights are normalized to have unit length before the dot product for reducing a statistical difference between the base and new classes. In contrast, we directly use sigmoid activation of the classification head for predicting the class of the bounding box, and train our fully-connected classifier with the focal loss \cite{lin2017focal}. Our \Approach~uses the sigmoid activation to predict each class independently, and thus alleviate the aforementioned bias.

% \textbf{Uncertainty Estimation in segmentation}: \cite{kendall2017uncertainties} introduces uncertainty estimation for semantic segmentation. There are two types of uncertainty: aleatoric and epistemic. The former captures noise inherent in the observations while the latter accounts for the uncertainty in the model parameters which can be reduced given enough data. \cite{kohl2018probabilistic} and \cite{kendall2015bayesian} are two examples capturing aleatoric uncertainty and epistemic uncertainty respectively. We introduce a new approach to model epistemic uncertainty in classification with an analytical formulation instead of Monte Carlo sampling as in prior work. Also, we leverage the estimated aleatoric uncertainty in box regression as a guidance for refining the initial box to better capture the object extent.

\section{Specification of \Approach}
\label{sec:approach}

% In this section, we first formally describe the problem statement as well as the overview of \Approach. Then the our contributions are discussed in detail subsequently.

\subsection{Problem Statement}
We address N-way K-shot iFSOD and iFSIS, where abundant training examples of $N_b$ base classes are provided for pre-training. After that, access to  training examples of the base classes is not available. When a few training examples $K$ of additional $N_n$ new classes are arbitrarily provided, our goal is to detect and segment all object instances belonging to all $N=N_b + N_n$ classes in a query image.

In the following, we specify our two contributions -- the 
probit classifier and uncertainty-guided box predictor.  

\subsection{The Probit Classifier}
As mentioned in Sec.~\ref{sec:related_work}, our \Approach~uses sigmoid activation instead of softmax activation for predicting the class of a bounding box. With sigmoid activation, the scores for all classes are independently predicted, and thus our classification head effectively addresses the statistical difference between the base and new classes. However, in the FSOD and FSIS settings, our experiments (see Tab.~\ref{tab:ablation_fsis}) demonstrate that the classification head with the sigmoid activation usually gives a lower performance than that with the softmax activation, when weights of the classifier head (also referred to as class weights) are learned as point estimation. 

To address this problem, we resort to Bayesian learning of a distribution of the class weights, and adopt the common variational framework. Formally, the class weights $w$ are characterized by the normal distribution, $w\sim \mathcal{N}(\mu,\Sigma)$, with mean $\mu \in \Real^{D}$ and diagonal covariance matrix $\Sigma \in \Real^{D \times D}_+$. Our goal is to learn $\mu$ and $\Sigma$ by minimizing the following variational objective:
\begin{align}
    L_c = l_d (p(c | f, \mu,\Sigma), c^*) + \text{KL}(\mathcal{N}(\mu,\Sigma) || \mathcal{N}(\textbf{0},\textbf{1})),
\label{eq:class_loss}
\end{align}
where $f \in  \Real^{D}$ is a feature extracted from the bounding box; $c$ and $c^*$ are the predicted and ground-truth classes; $l_d$ is the incurred sigmoid focal loss \cite{lin2017focal} for predicting $c$;  KL stands for the Kullback–Leibler divergence; and $p(c | f, \mu, \Sigma)$ is the posterior predictive distribution defined as
\begin{align}
    p(c| f, \mu, \Sigma) 
    & = \int \sigma(f^\top w) \mathcal{N}(w | \mu,\Sigma) dw,
    \label{eq:integral}
\end{align}
where $\sigma(\cdot)$ denotes the sigmoid activation.
% , and ``$\propto$'' means proportional up to a normalization constant. 

Once $\mu$ and $\Sigma$  are learned, class prediction amounts to the MAP problem: $c = \argmax_{c'}p(c' | f, \mu, \Sigma)$. %
 However, the integral in \eqref{eq:integral} is intractable. Prior work typically approximates the MAP problem with Monte Carlo sampling: $p(c| f, \mu, \Sigma) \approx \frac{1}{T} \sum_{w_t \sim \mathcal{N}(\mu,\Sigma)}{\sigma(f^\top w_t)}$
where $T$ is the number of Monte-Carlo samples. The Monte Carlo approximation, however, has a poor trade off between efficiency and accuracy, where using a sufficiently large $T$ would make our training and testing prohibitively slow.

Instead of the stochastic Monte Carlo sampling, we specify a more efficient deterministic approximation of the posterior predictive distribution. We first observe that our fine-tuning of the classification head's last layer using Bayesian learning is equivalent to learning a {\em Bayesian logistic regression} (BLR). Conveniently, the well-known probit function $\Phi(x)$  \cite{spiegelhalter1990sequential}, \cite[p.~219]{bishop:2006:PRML} provides a deterministic approximation to BLR. The probit function approximates the sigmoid function as $\sigma(x)\approx\Phi(\lambda x)=\frac{1}{2}\left[1+\text{erf}\left(\frac{\lambda x}{\sqrt{2}}\right)\right]$, where $\text{erf}(\cdot)$ is the error function, and $\lambda^2=\pi/8$ ensures that the two functions have the same slope at the origin. An important property of the probit function is that its convolution with a Gaussian function can be expressed analytically. Let $a = f^\top w \in \Real$ be a random variable whose expectation and variance can be expressed as $\mathbb{E}[a] = \mu_a = f^\top \mu \in \Real$ and  $\mathbb{V}[a]= \Sigma_a = f^\top \Sigma f \in \Real$. Then the posterior predictive distribution given by \eqref{eq:integral} can be efficiently approximated as 
\begin{align}
    & p(c| f, \mu, \Sigma) = \int \sigma(a) \mathcal{N}(a | \mu_a,\Sigma_a) da, \label{eq:before}\\
    & \approx \int \Phi(\lambda a) \mathcal{N}(a | \mu_a,\Sigma_a) da =  \Phi \left( \frac{\lambda \mu_a}{(1 + \lambda^{2}\Sigma_a)^{\frac{1}{2}}} \right) ,\\
      &\approx \sigma\left(\frac{f^\top \mu}{(1+\frac{\pi}{8}f^\top \Sigma f)^{\frac{1}{2}}} \right).
    \label{eq:probit}
\end{align}
%
%For a detailed derivation of Eq.~\eqref{eq:before}, the interested reader is referred to
%\cite[p.~219]{bishop:2006:PRML}.

As our classifier head uses the probit function for the MAP class prediction, we call it the probit classifier. It is suitable for iFSOD and iFSIS for two reasons. It leverages Bayesian learning to address the paucity of training data. Also, it predicts a score for each class independently to address the incremental-learning setting.

\subsection{Uncertainty-Guided Bounding Box Predictor}

Object appearances, shapes, and scales in test images may significantly differ from a few training examples available. Also, target objects in query images may be subject to partial occlusion. All this gives rise to uncertainty in bounding box prediction. We seek to explicitly model this uncertainty when predicting four offset values  $\{m_k\}_{k=1..4}\in \Real^4$ that initially identify the location of bounding boxes. Specifically, as shown in Fig.~\ref{fig:main_diagram}, our box predictor additionally estimates four uncertainty values $\{u_k\}_{k=1..4}\in \Real^4_+$ of the bounding box prediction, one for each of the $\{m_k\}$ predictions. The estimated uncertainty $\{u_k\}$ is then used as input along with the ROI-align-pooled features -- extracted from the initially predicted box $m$ -- to the box refinement module for the final offset bounding-box prediction $\{b_k\}_{k=1..4}\in \Real^4$.

To learn how to predict uncertainty $\{u_k\}$, initial bounding-box  $\{m_k\}$, and refined bounding-box  $\{b_k\}$  on a few training examples, we specify the following box loss:
\begin{equation}
    L_b = L_{\text{u}} + L_{\text{refine}},
\label{eq:boxloss}
\end{equation}
where $L_{\text{u}}$ is our new uncertainty-weighted box loss and $L_{\text{refine}}$ is loss incurred by the box refinement module. 

We define $L_{\text{u}}$ as
\begin{equation}
    L_{\text{u}}= \sum_{k=1}^{4} \frac{1}{2} \left ( \frac{(m_k - b^*_k)^2}{u^2_k} + u^2_k \right ), 
\label{eq:box_uncertainty}
\end{equation}
where $b^*$ is the ground-truth box. The first term in \eqref{eq:box_uncertainty} is aimed at minimizing a weighted difference between the ground-truth and predicted boxes. The weighting is inversely proportional to the predicted uncertainty $u_k$ such that the lower loss is incurred for box predictions with high uncertainty. The second term in \eqref{eq:box_uncertainty} is aimed at minimizing uncertainty values such that the network incurs a penalty for predicting high uncertainty when trying to reduce the first term in \eqref{eq:box_uncertainty}. 

When the box refinement module makes the final prediction $b$, it incurs the following loss:
\begin{equation}
    L_{\text{refine}} = \sum_{k=1}^{4} \text{smooth L1} (b_k, b_k^*).
\label{eq:box_refine}
\end{equation}

It is worth noting that our loss formulation fundamentally differs from other recent approaches aimed at estimating uncertainty in object detection. For example, recent approaches \cite{he2019bounding,lee2020localization} make the assumption that the bounding-box location and its uncertainty are governed by a Gaussian distribution. In contrast, we do not explicitly specify any probability distribution of box locations. 
\Approach~appears related to Cascade-RCNN \cite{cai2018cascade,cai2019cascade} which also refines the initial box. However, these approaches do not explicitly predict uncertainty and thus cannot use uncertainty as an input feature for the box refinement as we do. By contrast, our uncertainty estimation in training serves to transfer ``knowledge'' from the base classes to the new testing classes. In experiments, these approaches show worse performance than our uncertainty-guided box refinement module.  

\subsection{Our Training and Testing Strategies}
\label{sec:strategy}
% We follow the training scheme introduced in \cite{wang2020frustratingly} which is illustrated in Fig.~\ref{fig:problem_statement}~(a).
% This section describes our training and testing strategies.
% \begin{enumerate}[itemsep=-1pt,topsep=1pt, leftmargin=20pt]
%     \item \underline{Training on the base classes}:
\noindent \textbf{Training on the base classes}
\begin{enumerate}[itemsep=-1pt,topsep=2pt]
%[itemsep=-1pt,topsep=1pt, leftmargin=0pt]
    \item Obtain a variant of Mask-RCNN named Mask+Sigmoid by replacing the standard softmax classifier with the sigmoid classifier. Train Mask+Sigmoid  with the following loss functions: sigmoid focal loss, $L_{\text{refine}}$ in \eqref{eq:box_refine}, and mask-BCE loss.

    \item Obtain a variant of Mask+Sigmoid named Mask+Sigmoid+Uncertainty by replacing the box predictor with the uncertainty-guided box predictor (contribution 2 in Fig.~\ref{fig:main_diagram}). While freezing other modules, train the uncertainty-guided box predictor of Mask+Sigmoid+Uncertainty with loss $L_b$ in \eqref{eq:boxloss}.
    
    \item Store the class weights $\mu_b$ of the base classes for the sigmoid classifier of Mask+Sigmoid+Uncertainty.
    
\end{enumerate}
    
    % \item \underline{Fine-tuning on the new classes}: 
\noindent \textbf{Fine-tuning on the new classes}
\begin{enumerate}[itemsep=-1pt,topsep=2pt]
    
    \item 
    Obtain a variant of Mask+Sigmoid+Uncertainty named \Approach~ by replacing the sigmoid classifier with the probit classifier (contribution 1 in Fig.~\ref{fig:main_diagram}). 
    \item 
    While freezing other modules, train the probit classifier with loss $L_c$ in \eqref{eq:class_loss} to obtain the class weights $\mu_n, \Sigma_n$ of the new classes. Also, train the last layer of the box predictor with $L_b$ in \eqref{eq:boxloss}; and the last layer of the segmentation head with the mask-BCE loss.
\end{enumerate}
    
    % \item \underline{Testing on the base and new classes}: 
\noindent \textbf{Testing on the base and new classes}
\begin{enumerate}[itemsep=-1pt,topsep=2pt]
    \item Set $\mu = [\mu_b; \mu_n]$ and $\Sigma = [\textbf{0}; \Sigma_n]$ for the probit classifier, where $[\cdot; \cdot]$ is a concatenation. Also, concatenate the weights estimated for the base and new classes to obtain the box and segmentation-mask heads.
    \item Run \Approach~on query images.
\end{enumerate}
    
% \end{enumerate}

% It is worth noting that, our training strategy can be easily extended to multiple batches of new classes arriving sequentially as in continual learning (CL) by running step 2 for each batch of the new classes. Finally, we aggregate the obtained weights for probit classifier as in step 3 and 

\begin{table*}[h!]
\begin{center}
\small
\setlength{\tabcolsep}{4pt}
\begin{tabular}{l|cccccc|cccccc}
\hline 
\hline
& \multicolumn{6}{c}{Object Detection} \vline & \multicolumn{6}{c}{Instance Segmentation} \\
\hline
Number of shots & 1 & 2 & 3 & 5 & 10 & 30 & 1 & 2 & 3 & 5 & 10 & 30  \\
 \hline
Mask-RCNN           & 3.58 & 5.07 & 5.79 & 7.81 & 8.59 & 12.68 & 3.71 & 5.24 & 5.29 & 7.66 & 8.46 & 11.09  \\
Mask+Cosine         & 3.39 & 4.87 & 5.19 & 6.87 & 7.96 & 12.52 & 3.40 & 5.00 & 4.75 & 6.68 & 7.72 & 11.03 \\
\hline
Mask+Sigmoid        & 3.60 & 4.49 & 5.63 & 7.06 & 7.68 & 11.40 & 3.92 & 4.63 & 5.63 & 7.15 & 7.67 & 10.94 \\
Mask+Probit         & \textcolor{blue}{5.18} & \textcolor{blue}{5.90} & \textcolor{blue}{7.82} & \textcolor{blue}{9.45} & \textcolor{blue}{10.43} & \textcolor{blue}{13.48} & \textcolor{blue}{5.15} & \textcolor{blue}{6.03} & \textcolor{blue}{7.67} & \textcolor{blue}{9.34} & \textcolor{blue}{9.52} & \textcolor{blue}{12.07} \\
Mask+MC             & 4.52 & 5.45 & 7.02 & 8.85 & 9.57 & 13.05 & 4.54 & 5.19 & 6.91 & 8.26 & 8.98 & 11.55 \\
Mask+Sig+Uncert     & 4.48 & 5.32 & 6.76 & 8.49 & 9.16 & 12.85 & 4.84 & 5.88 & 7.00 & 8.62 & 9.22 & 11.98 \\
Mask+Sig+Gauss      & 3.74 & 4.77 & 5.89 & 7.33 & 7.86 & 11.65 & 3.94 & 4.72 & 5.87 & 7.24 & 7.76 & 11.02 \\
Mask+Sig+Refine     & 3.87 & 4.57 & 5.78 & 7.48 & 8.23 & 11.95 & 3.99 & 4.77 & 5.68 & 7.40 & 7.87 & 11.01\\
\Approach           & \textcolor{red}{6.34} & \textcolor{red}{6.93} & \textcolor{red}{8.93} &  \textcolor{red}{10.53} & \textcolor{red}{11.27} & \textcolor{red}{14.66} &  \textcolor{red}{5.54} & \textcolor{red}{6.33} & \textcolor{red}{7.80} & \textcolor{red}{9.41} & \textcolor{red}{10.23} & \textcolor{red}{13.08}  \\
\hline
\hline
\end{tabular}
\end{center}
\vspace{-10pt}
\caption{Our ablation study on FSOD and FSIS with different $K=\{1,2, 3, 5,10, 30\}$ on COCO. The best results are in \textcolor{red}{red}, second-best results are in \textcolor{blue}{blue}. Mask-RCNN and Mask+Cosine use the softmax activation with cross-entropy loss for training. The remaining ablations use the sigmoid activation with the focal loss for training.}
\label{tab:ablation_fsis}
\vspace{-5pt}
\end{table*}

\begin{table*}[h!]
\begin{center}
\small
\setlength{\tabcolsep}{2pt}
\begin{tabular}{l|ccc|ccc|ccc|ccc|ccc|ccc}
\hline 
\hline
& \multicolumn{9}{c}{Object Detection} \vline & \multicolumn{9}{c}{Instance Segmentation} \\
\hline
Tested on & \multicolumn{3}{c}{New classes} \vline & \multicolumn{3}{c}{Base classes} \vline & \multicolumn{3}{c}{All classes} \vline & \multicolumn{3}{c}{New classes} \vline & \multicolumn{3}{c}{Base classes} \vline & \multicolumn{3}{c}{All classes} \\
\hline
Number of shots & 1 & 5 & 10 & 1 & 5 & 10 & 1 & 5 & 10 & 1 & 5 & 10 & 1 & 5 & 10 & 1 & 5 & 10 \\
 \hline
 
% gMask-RCNN       & 2.70 & 7.24 & 8.66 & 31.71 & 34.06 & 34.89 & 24.45 & 27.36 & 28.33 & 3.01 & 7.49 & 9.01 & 30.77 & 32.20 & 32.51 & 23.83 & 26.02 & 26.95 \\

% MTFA \cite{Ganea_2021_CVPR} & 2.10 & 6.22 & 8.28 & 31.73 & 33.11 & 33.83 & 24.32 & 26.39 & 27.44 & 2.34 & 6.38 & 8.36 & 29.85 & 31.29 & 31.84 & 22.98 & 25.07 & 25.97  \\

% \hline

TFA \cite{wang2020frustratingly} & 2.90 & 7.00 & 9.10 & 31.90 & 32.30 & 32.40 & 3.60 & 11.50 & 14.20 & - & - & - & - & - & - & - & - & - \\

FSDetView \cite{xiao2020few} & 3.35 & 8.53 & 12.50 & 25.75 & 25.05 & 24.82 & 20.15 & 20.92 & 21.74 & - & - & - & - & - & - & - & - & - \\

GIFSOD \cite{li2021towards} & - & - & 8.50 & - & - & 28.10 & - & - & 23.20 & - & - & - & - & - & - & - & - & -  \\

ONCE \cite{perez2020incremental} & 0.70 & 1.00 & 1.20 & 17.90 & 17.90 & 17.90 & 13.60 & 13.70 & 13.70 & - & - & - & - & - & - & - & - & -  \\

LEAST \cite{li2021class} & 4.40 & 9.40 & 12.50 & 24.60 & 25.20 & 23.10 & 7.50 & 13.70 & 16.20 & - & - & - & - & - & - & - & - & -  \\

iMTFA \cite{Ganea_2021_CVPR} & 3.23 & 6.07 & 6.97 & 27.81 & 24.13 & 23.36 & 21.67 & 19.62 & 19.26 & 2.81 & 5.19 & 5.88 & 25.90 & 22.56 & 21.87 & 20.13 & 18.22 & 17.87  \\

\hline

Mask+Sigmoid   & 2.85 & 6.34 & 8.04 & 38.55 & 38.53 & 38.53 & 29.62 & 30.49 & 30.91 & 3.06 & 6.52 & 8.00 & 35.70 & 35.69 & 35.69 & 27.54 & 28.76 & 29.37 \\

\Approach       & \textbf{4.54} & \textbf{9.91} & \textbf{12.55} & \textbf{40.08} & \textbf{40.06} & \textbf{40.05} & \textbf{31.19} & \textbf{32.52} & \textbf{33.02} & \textbf{3.95} & \textbf{8.80} & \textbf{10.06} & \textbf{36.35} & \textbf{36.33} & \textbf{36.32} & \textbf{28.45} & \textbf{29.89} & \textbf{30.41} \\
\hline
\hline
\end{tabular}
\end{center}
\vspace{-10pt}
\caption{iFSOD and iFSIS results on COCO with different $K=\{1, 5, 10\}$. `-' indicates no results are reported. 
% MFTA \cite{Ganea_2021_CVPR} is the modified version of TFA \cite{wang2020frustratingly} trained with gFSOD and gFSIS settings as described in Tab.~\ref{tab:settings}. 
Best results are in \textbf{bold}.}
\label{tab:coco_results}
\vspace{-10pt}
\end{table*}

\begin{table*}[h]
\begin{center}
\small
\setlength{\tabcolsep}{5pt}
\begin{tabular}{l|c|c|cccc|cccc}
\hline 
\hline
Settings & FSOD & FSIS & \multicolumn{4}{c}{Object Detection} \vline & \multicolumn{4}{c}{Instance Segmentation}  \\
\hline
Tested on & New & New & New & Base-c & Base-f & All & New & Base-c & Base-f & All \\
\hline
TFA \cite{wang2020frustratingly} (gFSOD)       & 18.35 & - & 16.90 & 24.30 & 27.90 & 24.40 & - & - & - & -  \\

Mask-RCNN \cite{he2017mask} (gFSOD \& gFSIS)      & 16.50 & 18.31 & 12.11 & 24.54 & 28.59 & 24.04 & 12.75 & 25.35 & 27.75 & 24.36  \\
\hline
% iMTFA \cite{Ganea_2021_CVPR} (iFSOD \& iFSIS) & 16.54 & 18.27 & 14.28 & 16.35 & 18.27 & 16.28 & 16.27 & 16.28 & 18.69 & 16.46  \\
Mask+Sigmoid (iFSOD \& iFSIS)     & 16.93 & 19.18 & 15.02 & 23.33 & 27.23 & 23.55 & 17.39 & 25.26 & 27.05 & 24.75  \\
\Approach~(iFSOD \& iFSIS)          & \textbf{20.76} & \textbf{21.06} & \textbf{18.38} & \textbf{26.11} & \textbf{30.12} & \textbf{26.46} & \textbf{18.26} & \textbf{26.29} & \textbf{28.46} & \textbf{25.90}  \\
% \Approach~(iFSIS, R101)        & \textcolor{red}{25.78} & \textcolor{red}{28.68} & \textcolor{red}{23.61} & \textcolor{red}{27.83} & \textcolor{red}{31.67} & \textcolor{red}{28.64} & \textcolor{red}{21.34} & \textcolor{red}{27.96} & \textcolor{red}{29.67} & \textcolor{red}{27.61} \\
\hline
\hline
\end{tabular}
\end{center}
\vspace{-10pt}
\caption{Object detection and instance segmentation results with AP metric for FSOD, FSIS, iFSOD, and iFSIS tasks on LVIS. The best results are in \textbf{bold}. TFA and Mask-RCNN are trained with more supervision than Mask+Sigmoid and \Approach~as described in Tab.~\ref{tab:settings}. 
% R50 and R101 indicate the backbone networks are ResNet50 and ResNet101 respectively.
Base-c and Base-f indicate the common classes ($\ge 100$ images) and frequent classes (10-100 images) among the base classes.}
\label{tab:lvis_results}
\vspace{-10pt}
\end{table*}

\section{Experimental Results}
\label{sec:experiments}

\textbf{Datasets \& Metrics}: We evaluate \Approach~on the modified version of the COCO 2014 dataset \cite{lin2014microsoft} introduced by \cite{kang2019few} for FSIS and FSOD. Also, we are the first to evaluate iFSOD, FSIS, and iFSIS on a new split of the LVIS dataset \cite{gupta2019lvis} introduced by \cite{wang2020frustratingly} for FSOD. We report the common COCO-style evaluation metrics of both object detection and instance segmentation  -- namely, the average precision (AP) at multiple intersection-over-union (IoU) thresholds ranging from 0.5 to 0.95.

For COCO, the 20 categories shared with PASCAL VOC \cite{Everingham10} are used as new classes while the remaining 60 classes are used as the base classes. We vary the number of examples for the new classes, i.e. $K=\{1, 2, 3, 5, 10, 30\}$, and report average results with $95\%$ confidence interval over 10 runs with different sets of few-shot examples for each $K$. This paper reports results for $K=\{1, 5, 10\}$ for brevity. Other results are in the supplementary material.
% The AP-mask results are reported in this paper while the AP-box results are reported in the supplementary material.

LVIS has 1230 classes where some have a large number of examples and some other, called rare classes, have only a few examples (less than 10 examples per class). Hence, the number of images for each class in LVIS  has a long-tail distribution. We take the frequent (appearing in more than 100 images) classes and the common  classes (10-100 images) in LVIS as the base classes, while the 454 rare classes (appearing less than 10 images) as the new classes. Due to the small number of training examples for the rare classes in LVIS, we cannot have multiple runs as in COCO, thus we follow \cite{wang2020frustratingly}  on their split with $K \leq 10$.

\subsection{Implementation Details}

Our backbone CNN is ResNet-50 \cite{he2015deep} with the FPN of \cite{lin2017feature}, as in recent work on FSIS. All variants of \Approach~are implemented using the detectron2 toolbox \cite{wu2019detectron2} with the codebase of \cite{wang2020frustratingly}. All variants of our approach, specified in Sec.~\ref{sec:ablation_study}, are trained using SGD and a batch size of 16 on 8 NVIDIA GPUs V100s, with two images per GPU. The learning rate is set to 0.02 and 0.01 for the pre-training and fine-tuning stages respectively. The number of iterations for the pre-training stage is 110000 with two weight decay steps with the rate of 10 at 80000 and 100000 iterations. The number of iterations for fine-tuning stage depends on the number of examples ranging from 500 iterations (with $K=1$) to 6000 iterations (with $K=30$). The hyper-parameters of the focal loss are $\gamma=0.25, \alpha=2$.

The threshold for filtering predictions before the NMS is 0.05. To select proposals for training the uncertainty-guided box predictor, we choose the predicted boxes with IoU larger than 0.7 with their closest ground-truth boxes. The threshold for deciding foreground and background in the segmentation mask head is 0.5. We use the softplus function $f(x) = \ln(1 + \exp(x))$ to ensure the class variances $\Sigma$ and box uncertainty $u$ are non-negative.

Our segmentation head is the same as in Mask-RCNN \cite{he2017mask}. It is trained with the binary cross-entropy (BCE) loss. Our box refinement module has the same architecture as the box head. The final layers of the uncertainty-guided box predictor and segmentation head are class-specific, and obtained by concatenating the class weights learned for the base and new classes.

% \begin{figure*}[h!]
%     \centering
%     \includegraphics[scale=0.43]{images/compare_with_SOTA_iFSIS_mask_new.pdf}
%     % \vspace{-10pt}
%     \caption{Average AP metric with 95\% confidence interval over 10 runs on iFSIS on COCO. $K=\{1,2,3,5,10,30\}$ are the number of few-shot examples used in the fine-tuning stage.}
%     \label{fig:compare_sota_iFSIS}
%     % \vspace{-10pt}
% \end{figure*}

% \begin{figure}[h!]
%     \centering
%     \includegraphics[scale=0.5]{images/compare_with_SOTA_FSIS_mask_new.pdf}
%     % \vspace{-10pt}
%     \caption{Average AP metric with 95\% confidence interval over 10 runs on FSIS (test on the new classes only) on COCO. $K=\{1,2,3,5,10,30\}$ are the number of few-shot examples used in the fine-tuning stage.}
%     \label{fig:compare_sota_FSIS}
%     \vspace{-7pt}
% \end{figure}

\subsection{Ablation Study}
\label{sec:ablation_study}

The following ablations are evaluated on the first run of COCO for studying how each component of \Approach~affects the final performance.
\begin{itemize}[itemsep=-1pt,topsep=1pt, leftmargin=10pt]
    \setlength\itemsep{-0.25em}
    \item \textit{Mask-RCNN}: the original Mask-RCNN \cite{he2017mask} with the softmax classifier.
    \item \textit{Mask+Cosine}: replace the dot product with the  cosine-similarity classifier in Mask-RCNN as in \cite{Ganea_2021_CVPR}.
    \item \textit{Mask+Sigmoid}: replace the softmax classifier with the sigmoid classifier in Mask-RCNN  (our strong baseline).
    \item \textit{Mask+Probit}: replace the sigmoid classifier with the probit classifier in Mask+Sigmoid, our contribution 1 in Fig.~\ref{fig:main_diagram}.
    \item \textit{Mask+MC}: replace the probit approximation with the Monte Carlo (MC) sampling in Mask+Probit, where the number of samplings is $T=10$.
    \item \textit{Mask+Sig+Uncert}: replace the box predictor in Mask+Sigmoid with the uncertainty-guided box predictor, our contribution 2 in Fig.~\ref{fig:main_diagram}.
    \item \textit{Mask+Sig+Gauss}: a variant of Mask+Sigmoid which additionally predicts box uncertainty (similar to \cite{he2019bounding}) with Gaussian distribution assumption.
    \item \textit{Mask+Sig+Refine}: similar to Cascade RCNN \cite{cai2018cascade} as it does not explicitly predict uncertainty when refining the initial box prediction
    \item Our \textit{\Approach}~in Fig.~\ref{fig:main_diagram}.
\end{itemize}

Tab.~\ref{tab:ablation_fsis} shows our evaluation of the above ablations on FSOD and FSIS (see Tab.~\ref{tab:settings}). 
From Tab.~\ref{tab:ablation_fsis}, Mask-RCNN with the softmax activation outperforms other point-estimation-based ablations with the sigmoid activation. However, Mask-RCNN gives worse performance than the ablations which use the sigmoid activation with Bayesian learning for the Probit or MC. This justifies our choice of Bayesian-based classifiers with the sigmoid activation.
Also, our probit classifier outperforms the MC classifier, suggesting that MC requires  longer stochastic sampling $T\gg 10$, which would be prohibitively slow in practice. Importantly, our modeling of uncertainty in Mask+Sig+Uncert gives a  substantial performance gain over Mask+Sigmoid, Mask+Sig+Gauss, and Mask+Sig+Refine. This justifies our contribution 2. Finally, our full approach, \Approach~yields the best performance, about +2.5 performance gain over the strong baseline Mask+Sigmoid and +2 over Mask-RCNN. In the following experiments, we choose Mask+Sigmoid and \Approach~ to compare with prior work.

% Fig.~\ref{fig:ablation_iFSIS} compares Mask-RCNN, Mask+Sigmoid, and \Approach~on iFSIS, as well as Mask-RCNN on gFSIS (see Tab.~\ref{tab:settings}). Note that Mask-RCNN in the gFSIS setting gets more supervision than in the iFSIS setting, since fine-tuning in gFSIS is allowed access to training examples of the base classes, whereas iFSIS prohibits this access. From Fig.~\ref{fig:ablation_iFSIS}, Mask-RCNN (iFSIS) produces bad performance on the new classes, as it tends to predict the base classes all the time, in comparison with  Mask+Sigmoid (iFSIS) and \Approach~(iFSIS). This justifies our choice of the sigmoid activation instead of the softmax one in Mask-RCNN. Interestingly, \Approach~(iFSIS) significantly outperforms Mask-RCNN (gFSIS) on the new and base classes, despite the fact that Mask-RCNN (gFSIS) has access to more supervision. Importantly, the results demonstrate that \Approach~retains good performance on the base classes.

% Additional evaluations of the ablations for the task of object detection in the FSIS and iFSIS settings are reported in the supplementary material.

\subsection{Comparison with Prior Work on COCO}

% Fig.~\ref{fig:compare_sota_iFSIS} (iFSIS) and Fig.~\ref{fig:compare_sota_FSIS} (FSIS) 

Tab.~\ref{tab:coco_results} compares our results on COCO in iFSOD and iFSIS with the strong baseline Mask+Sigmoid, approaches designed for FSOD and FSIS (TFA \cite{wang2020frustratingly}, FSDetView \cite{xiao2020few}), and approaches designed for iFSOD and iFSIS (GIFSOD \cite{li2021towards}, ONCE \cite{perez2020incremental}, LEAST \cite{li2021class}, MTFA, and iMTFA \cite{Ganea_2021_CVPR}). 

% As mentioned in Sec. \ref{sec:related_work}, there is a scant work on FSIS including Meta-RCNN \cite{yan2019meta}, Siamese-Mask-RCNN \cite{michaelis2018oneshot}, FGN \cite{fan2020fgn}, and FAPIS \cite{nguyen2021fapis}. Siamese-Mask-RCNN and FAPIS are designed for 1-way FSIS, which is completely different from our N-way FSIS, so it is not trivial to apply to N-way iFSIS. Also, FGN has not released the code yet. Meta-RCNN is a variant of Mask-RCNN, where the feature vector representing a target class is extracted from few-shot examples and channel-wise multiplied with feature map of the query image to filter out irrelevant features. 

TFA (a fine-tuning-based approach) and FSDetView (an episodic-training-based approach) are adapted to the iFSOD setting as follows. TFA is first trained on the base classes, resulting in model 1. Then, TFA is fine-tuned on the new classes, resulting in model 2. Finally, we run the models 1 and 2 on the same test images and select the top 100 predictions, as in the COCO evaluation protocol. 
For FSDetView, after pre-training the model on training examples of the base classes, we use the pretrained model to estimate the prototype for each base class, and then run the pretrained model on the new classes to extract their prototypes. FSDetView uses the prototypes of both new and base classes for object detection in test images. As TFA and FSDetView were not originally designed for iFSOD, their aforementioned adaptation from FSOD to iFSOD has two limitations: (1) Storage and running of two distinct models/prototypes for predicting the base and new classes; (2) The two distinct models may independently each yield a high score on a base and new class in the test image, which is then hard to resolve.

Among the approaches aimed at iFSOD, ONCE is based on YOLO \cite{redmon2018yolov3}, and LEAST and GIFSOD are based on Faster-RCNN \cite{ren2015faster}. iMTFA is a variant of Mask+Cosine, where the weight $\mu_n$ is set to the box feature $f$ extracted from an example of the new class. 

For the comparison in Tab.~\ref{tab:coco_results},  results are averaged over 10 runs with different sets of few-shot examples. From this table, for the iFSOD setting, \Approach~outperforms state of the art (SOTA) approaches slightly on new classes (+0.05) while significantly on the base classes (+12) for $K=10$. For the iFSIS setting, \Approach~outperforms the SOTA iMTFA with significant margins. Specifically, for $K=10$, our performance gains are about +6 for the new classes and +16 for the base classes on segmentation. Our performance gains are very large, especially for the challenging COCO dataset.

% Additional results for the task of object detection in the FSIS and iFSIS settings are reported in the supplementary material.

\begin{figure*}[h!]
    \centering
    \includegraphics[scale=0.55]{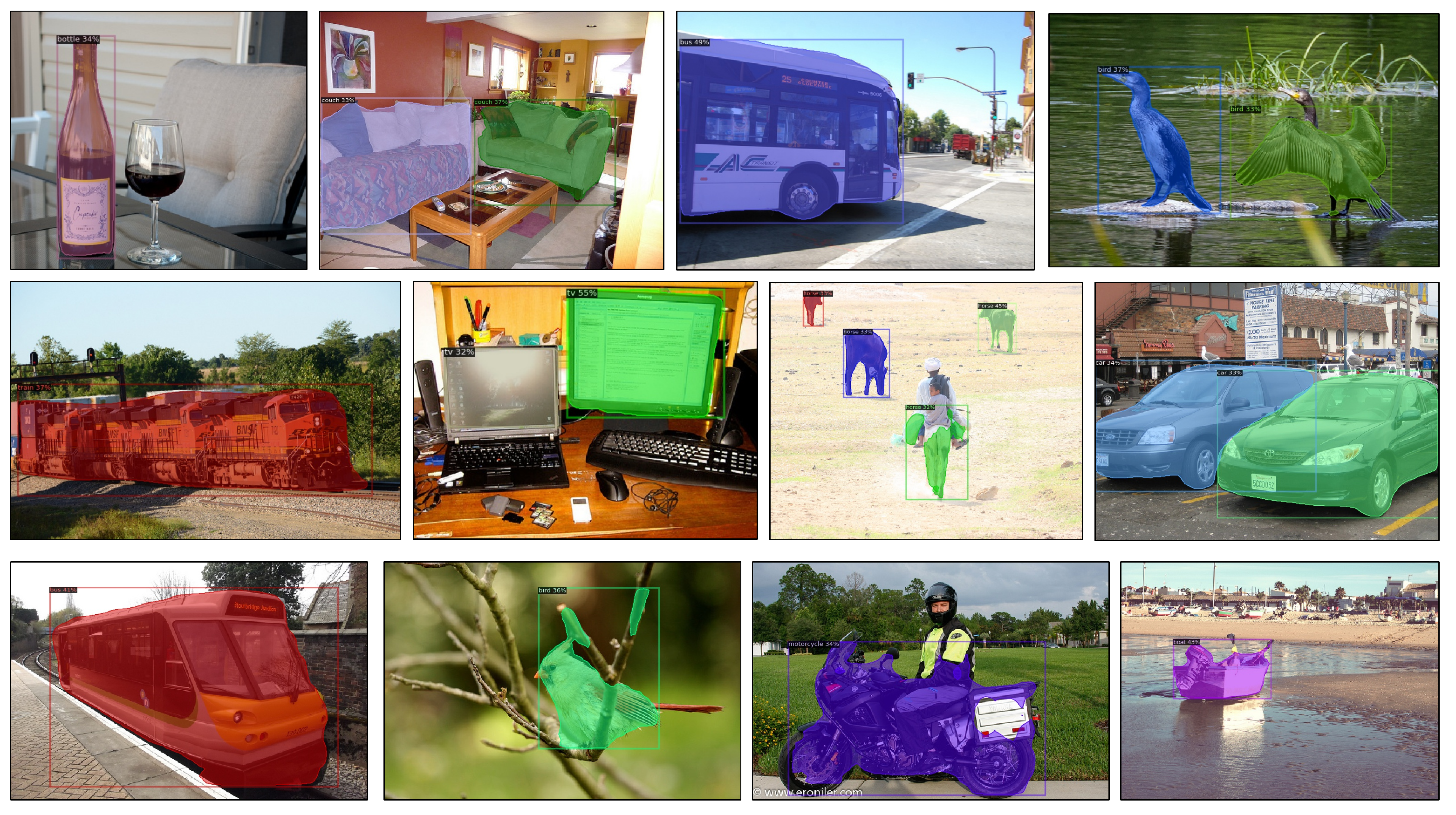}
    % \vspace{-10pt}
    \caption{Representative results of \Approach~on COCO-new with $K=30$. The top two rows show success cases while the bottom row shows failure cases. In the last row, from left to right: the train is detected as bus, bonding-box is too large for the bird, the human leg is segmented as a part of the motorbike, the small boats far behind are not detected.}
    \label{fig:general_qualitative}
    \vspace{-10pt}
\end{figure*}

\begin{figure}[h!]
    \centering
    \includegraphics[scale=0.34]{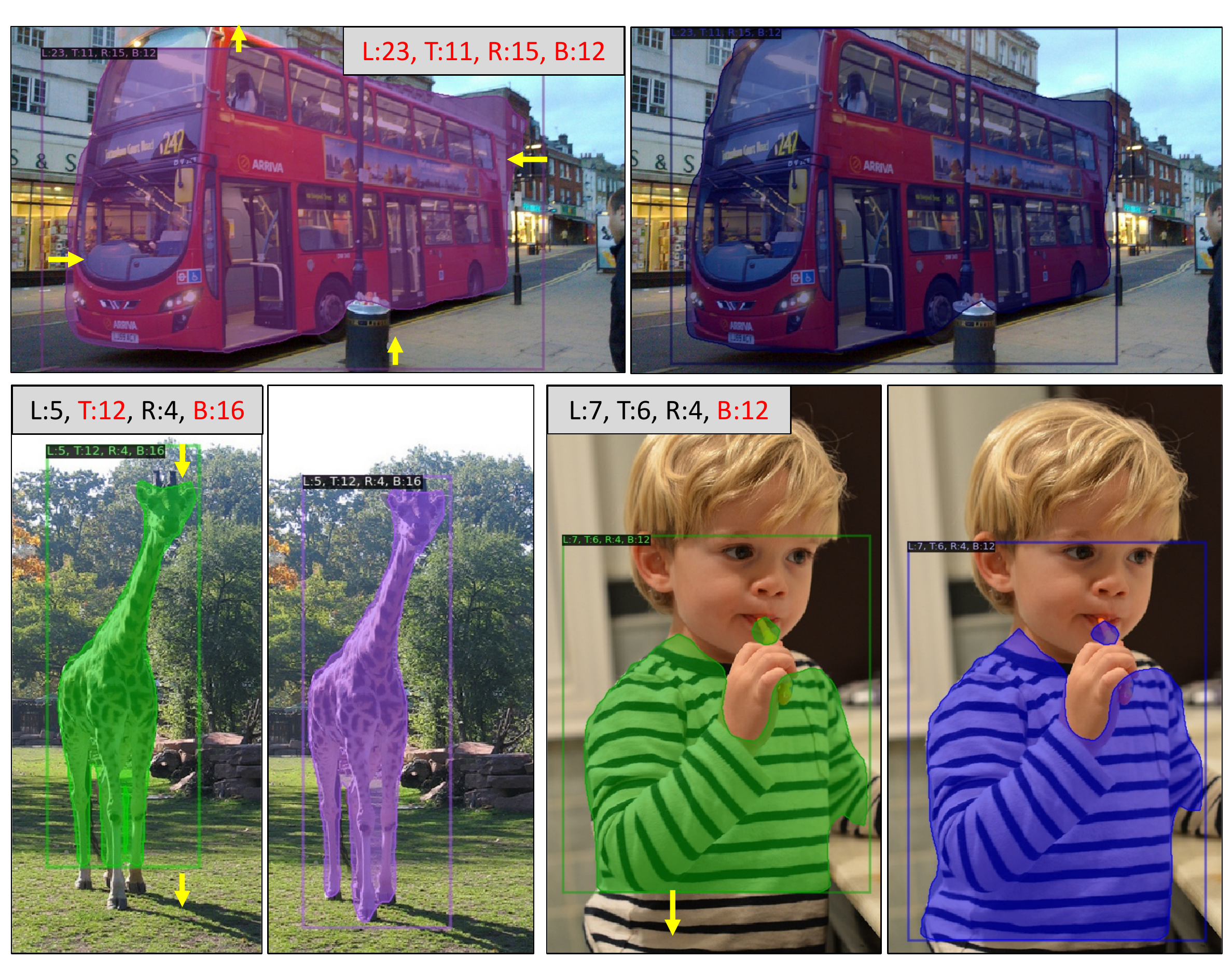}
    % \vspace{-10pt}
    \caption{Bounding-box refinement with our uncertainty-guided box predictor on COCO-new with $K=1$. For each pair, left: initial box, right: refined box. The box labels show left (L), top (T), right (R), and bottom (B) side uncertainties; red text indicates high uncertainty ($\geq 10$)   that lead to large refinements. Yellow arrows indicate the refined direction of each box side. }
    \label{fig:box_qualitative}
    \vspace{-10pt}
\end{figure}

\subsection{Results on LVIS}

Tab.~\ref{tab:lvis_results} reports our results on  LVIS with $K \leq 10$ on iFSIS and FSIS. Although Mask+Sigmoid uses less supervision than Mask-RCNN and TFA, they give comparable results. \Approach~significantly outperforms Mask-RCNN with the gains of +3 on FSIS and +6 on iFSIS on the new classes. These results demonstrate the effectiveness of our \Approach~on the more challenging dataset LVIS.

\subsection{Qualitative Evaluation}
 Fig.~\ref{fig:general_qualitative} illustrates some of our results. The top two rows show success cases while the bottom row shows failure cases. For the failure cases from left to right: the train is misclassified as bus due to very similar appearance, the bird detection has a large bounding box due to the occlusion, the human leg is segmented as a part of the motorbike due to similar appearance, and the small boats far behind are not detected. Fig.~\ref{fig:box_qualitative} shows our box refinement results. As can be seen, high uncertainty about the box prediction gives a considerable box refinement.

\section{Conclusion and Discussion}
\label{sec:conclusion}

We have specified \Approach~to address N-way K-shot incremental few-shot object detection (iFSOD) and instance segmentation (iFSIS). \Approach~leverages Mask-RCNN, but modifies the standard softmax classifier and bounding-box prediction head. The  softmax classifier has been replaced with the sigmoid classifier for alleviating a statistical imbalance of the base and new classes. The sigmoid classifier has been further improved via Bayesian learning to robustly estimate a distribution of the classifier head's weights on a few examples of the new classes. For this Bayesian learning, we have proposed an analytical approximation using the probit function. Also, the standard box predictor of Mask-RCNN has been extended to explicitly predict uncertainty of box prediction, and use the estimated uncertainty as input to a bounding-box refinement module. 
\Approach~significantly outperforms iMTFA -- the state of the art in iFSIS -- with very large performance gains on the COCO dataset, +6 on the new classes and +16 on the base classes with 10 training examples. Moreover, we are the first to report the results on the LVIS dataset for the iFSOD, FSIS, and iFSIS problems, where we outperform strong baselines.

% IMPORTANT!!!!!!!!
% New from this year, authors are encouraged to have a discussion on the limitations and potential negative societal impact 
%

As for potential limitations, our contribution 1, in general, could be applied to any object detector, including RPN-based ones like Faster-RCNN \cite{ren2015faster} and Mask-RCNN \cite{he2017mask}), point-based ones like FCOS \cite{tian2019fcos} and Center-Net \cite{zhou2019objects}), and transformer-based ones like DETR \cite{carion2020end}. However, in our experiments, we have succeeded in applying it only to the RPN-based detector. This might be due to the pre-selected balanced training examples of background and foreground classes in the second stage of Faster-RCNN, which is absent in the other two detector frameworks. %A study on how to apply our contribution 1 to other detector frameworks would be an interesting research direction. 
Also, while our estimation of uncertainty of box prediction and our new uncertainty-weighted box loss have enabled significant performance gains, our contribution 2  lacks a theoretical underpinning as to why our formulation outperforms a Gaussian-based uncertainty estimation.

%Our work is aimed at fundamental contributions to object detection and segmentation, and thus the extent of its negative societal impact goes together with that of the entire research area. 

As any system for object detection and segmentation, ours could be weaponized and misused for malicious violations of privacy, as well as used for efficiently discovering and fighting against such misuses due to its few-shot learning capabilities.

\textbf{Acknowledgement.} This work was supported in part by DARPA MCS Award N66001-19-2-4035.

%on privacy consideration, i.e., human tracking which heavily relies on the object detection results. A pretrained human detector on a group of people can be easily extended to other people with a few annotated examples. This could results in malicious tracking of people in surveillance camera. To mitigate this, protecting the pretrained model from unintended use should be a high priority. 

 %In FSIS, the classification and bounding-box heads of Mask-RCNN  are typically learned in a point-estimation fashion, and hence may poorly generalize to query images showing both the base and new classes under large  appearance-shape-scale variations. 

\onecolumn
\section{Appendix}

\textbf{Comparison to prior work on COCO with AP for object detection and instance segmentation}

\begin{figure*}[ht!]
    \centering
    \includegraphics[scale=0.41]{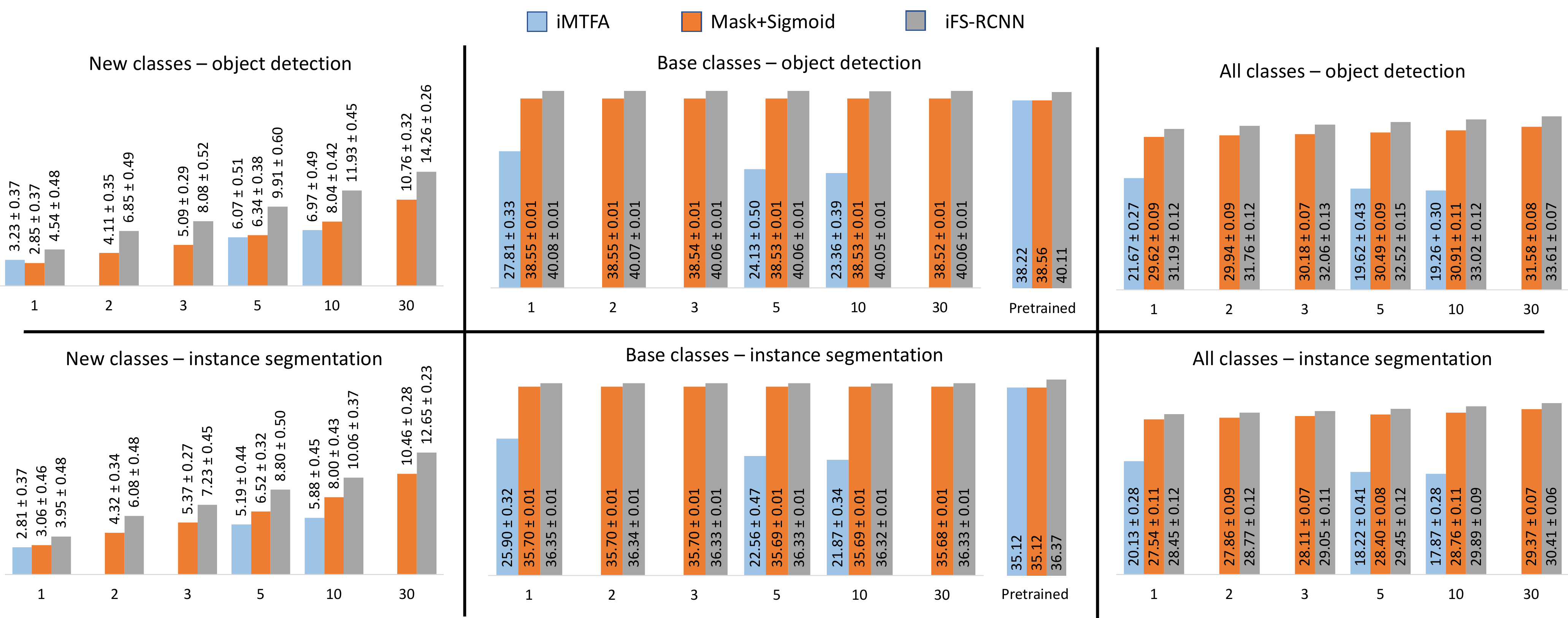}
    % \vspace{-10pt}
    \caption{Average AP metric with 95\% confidence interval over 10 runs on iFSIS task. $K=\{1,2,3,5,10,30\}$ are the number of few-shot examples used in the fine-tuning stage. Our baseline Mask+Sigmoid outperforms the state-of-the-art approach iMTFA in almost all metrics with significant margins (except the object detection with one shot on the new classes). Our final approach \Approach~even further improves that performance of Mask+Sigmoid. In base classes results, our baseline and our \Approach~ almost keep the performance of the pretrained models. This justifies the superior performance of sigmoid over softmax activation function for iFSOD and iFSIS.}
    \label{fig:compare_sota_iFSIS}
\end{figure*}

\textbf{Comparison to prior work on COCO with AP50 for object detection and instance segmentation}

\begin{figure*}[ht!]
    \centering
    \includegraphics[scale=0.43]{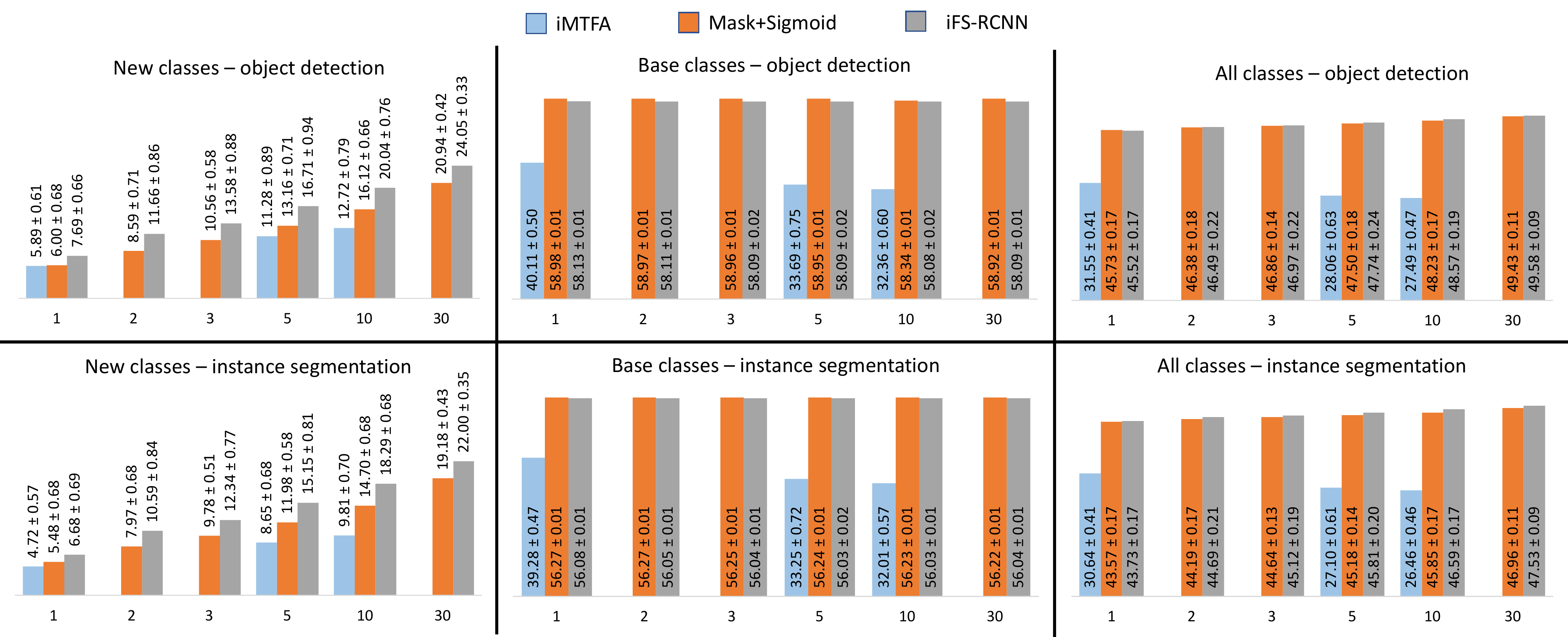}
    % \vspace{-10pt}
    \caption{Average AP50 metric with 95\% confidence interval over 10 runs on iFSIS task. $K=\{1,2,3,5,10,30\}$ are the number of few-shot examples used in the fine-tuning stage. The same trend can be observed as in the AP metric. See the caption of Fig.~\ref{fig:compare_sota_iFSIS}}
    \label{fig:compare_sota_iFSIS_AP50}
\end{figure*}
 
\twocolumn

%%%%%%%%% REFERENCES
{\small
\bibliographystyle{ieee_fullname}
\bibliography{egbib}
}

\end{document}